\documentclass{article}

\usepackage{iclr2022_conference,times}

\usepackage[utf8]{inputenc} 
\usepackage[T1]{fontenc}    
\usepackage{hyperref}       
\usepackage{url}            
\usepackage{booktabs}       
\usepackage{amsfonts}       
\usepackage{nicefrac}       
\usepackage{microtype}      
\usepackage{xcolor}         
\usepackage{CJKutf8}
\usepackage{graphicx}
\usepackage[frozencache,cachedir=.]{minted} 
\usepackage{multirow}
\hypersetup{
colorlinks=true,
linkcolor=black
}

\title{A Challenge on Semi-Supervised and Reinforced Task-Oriented Dialog Systems}

\author{Zhijian Ou\\Tsinghua University, Beijing, China
\And
Junlan Feng\\China Mobile, Beijing, China
\And
Juanzi Li\\Tsinghua University, Beijing, China
\And
Yakun Li\\ Tsinghua University, Beijing, China
\AND
Hong Liu\\Tsinghua University, Beijing, China
\AND
Hao Peng\\Tsinghua University, Beijing, China
\AND
Yi Huang\\China Mobile, Beijing, China
\AND
Jiangjiang Zhao\\China Mobile, Beijing, China
}
\author{
\begin{tabular}[t]{ll}
   \textbf{Zhijian Ou}  & Junlan Feng \\
    Tsinghua University, Beijing, China & China Mobile, Beijing, China\\
    &\\
    Juanzi Li & Yakun Li\thanks{Equal Contribution}\\
    Tsinghua University, Beijing, China & Tsinghua University, Beijing, China\\
    &\\
    Hong Liu$^{*}$ & Hao Peng$^{*}$\\
    Tsinghua University, Beijing, China & Tsinghua University, Beijing, China\\
    &\\
    Yi Huang & Jiangjiang Zhao\\
    China Mobile, Beijing, China & China Mobile, Beijing, China\\
\end{tabular}
}
\iclrfinalcopy 
\begin{document}

\maketitle

\tableofcontents
\renewcommand*\contentsname{\hfill Contents \hfill}

\onecolumn
\newpage
\section{Introduction}
Task-oriented dialogue (TOD) systems are designed to assist users to accomplish their goals, and have gained more and more attention in both academia and industry with recent advances in neural approaches \citep{williams2016dialog,gao2019neural}.
A TOD system typically consists of several modules, which track user goals to update dialog states, query a task-related knowledge base (KB) using the dialog states, decide actions and generate responses.
Unfortunately, building TOD systems remains a label-intensive, time-consuming task for two main reasons.
First, training neural TOD systems requires manually labeled dialog states and system acts (if used), in both traditional modular approach \citep{young2013pomdp,mrkvsic2017neural} and recent end-to-end trainable approach \citep{wen2017a, liu2017end, lei2018sequicity, fsdm, zhang2020task, liu2022mga}.
Second, it is often assumed that a task-related knowledge base is available. But for system development from scratch in many real-world tasks, expert labors are needed to construct the KB from annotating unstructured data.
Thus, the labeled-data scarcity hinders efficient development of TOD systems at scale.

Remarkably, unlabeled data are often easily available in many forms such as human-to-human dialogs, open-domain text corpus, and unstructured knowledge documents.
This has motivated the development of semi-supervised learning (SSL) \citep{zhu2006semi}, which aims to leverage both labeled and unlabeled data, for both information extraction to construct the knowledge base and building the TOD system itself.
Additionally, although it has long been recognized that TOD systems could be formulated as Markov Decision Processes (MDPs) and trained via reinforcement learning (RL) for policy learning for the agent \citep{young2013pomdp}, it remains very challenging to build reinforced TOD systems due to large language action spaces.
There are significant individual research threads, including semi-supervised information extraction \citep{li2019semi,song2020upgrading}, using pre-trained language models \citep{hosseini2020simple,peng2020etal} or latent variable models \citep{zhang2020probabilistic,liu2021variational} for semi-supervised TOD systems, grounded response generation with unstructured knowledge sources \citep{kim2020beyond}, reinforcement training of the system from interactions with user simulators \citep{kreyssig-etal-2018-neural,shi-etal-2019-build}, and so on.

The purpose of this challenge is to invite researchers from both academia and industry to share their perspectives on building \underline{se}mi-supervised and \underline{re}inforced \underline{TOD} systems and to advance the field in joint effort.
Hence, we refer to this challenge as the sereTOD challenge.
A shared task is organized for benchmarking and stimulating relevant researches. For the first sereTOD challenge, a large-scale TOD dataset is newly released, consisting of 100,000 real-world dialogs, where only 10,000 dialogs are annotated.

The remainder of this description document for the challenge is organized as follows. We first briefly describe techniques of interest for the sereTOD challenge. Then, we elaborate the shared task. The dataset is introduced in detail, including the structure of annotations and the guideline for annotations. Finally, the challenge rules is described.

\section{Techniques of Interest}
\label{sec:techniques}
This challenge encourages submissions on building semi-supervised and reinforced TOD systems. All types of semi-supervised techniques are welcome, such as, to name a few, pre-training, self-training, self-supervised, weakly-supervised, transfer learning for zero-shot or few-shots, latent-variable modeling, domain adaptation, and data augmentation. 
Both online and offline RL techniques are welcome.

Possible techniques include, but are not limited to, the following:
\begin{itemize}
    \item General techniques for task-oriented dialog systems
    \item Semi-supervised information extraction and knowledge modeling
    \item Grounded dialog with unstructured knowledge sources
    \item Semi-supervised task-oriented dialog systems
    \item Reinforced task-oriented dialog systems
    \item User simulators
\end{itemize}

\begin{table}[t]
		\caption{Comparison of our MobileCS corpus to MultiWOZ
		}
		\centering
		\resizebox{0.8\linewidth}{!}{
			\begin{tabular}{c c cc}
				\toprule
				\multirow{2}{*}{\textbf{Metric}} &\multirow{2}{*}{\textbf{MultiWOZ}}
	       	&\multicolumn{2}{c}{\textbf{MobileCS}}\\
				\cmidrule(lr){3-4}
				& &\textbf{labeled}  &\textbf{unlabeled} \\
				\midrule
				Dialogs  &8,438 &8,975 &87,933 \\
				Turns &113,556     &100,139 &972,573 \\
				Tokens  &1,490,615  & 3,991,197  &39,491,883 \\
				Avg. turns per dialog  &13.46 &11.16 &11.06 \\
				Avg. tokens per turn &13.13 &39.86 &40.61 \\
				Slots  &24  & 26   & -\\
				Values  &4,510  & 14,623  & -\\
         \bottomrule	
		\end{tabular}}
		\vspace{-1.0em}
		\label{tab:mwz-compare}
	\end{table}
	
\section{Shared Task}


We introduce a new shared task, aiming to benchmark semi-supervised and reinforced task-oriented dialog systems, built for automated customer-service for mobile operators. The task consists of two tracks: 
\begin{itemize}
    \item Information extraction from dialog transcripts (Track 1)
    \item Task-oriented dialog systems (Track 2)
\end{itemize}

An important feature for this shared task is that we release around 100,000 dialogs (in Chinese), which come from real-world dialog transcripts between real users and customer-service staffs from China Mobile, with privacy information anonymized. 
We call this dataset as MobileCS (mobile customer-service) dialog dataset, which differs from existing TOD datasets in both \emph{size} and \emph{nature} significantly.
To the best of our knowledge, MobileCS is not only the largest publicly available TOD dataset, but also consists of real-life data (namely collected in real-world scenarios). For comparison, the widely used MultiWOZ dataset consists of 10,000 dialogs and is in fact simulated data (namely collected in a Wizard-of-Oz simulated game).
See data statistics shown in Table \ref{tab:mwz-compare}.

A schema is provided, based on which 10,000 dialogs are labeled by crowdsourcing. The remaining 90,000 dialogs are unlabeled.
The teams are required to use this mix of labeled and unlabeled data to train information extraction models (Track 1), which could provide a knowledge base for Track 2, and train TOD systems (Track 2), which could work as customer-service bots.
We put aside 1,000 dialogs as evaluation data.


\begin{figure}[t]
    \centering
    \includegraphics[width=0.98\linewidth]{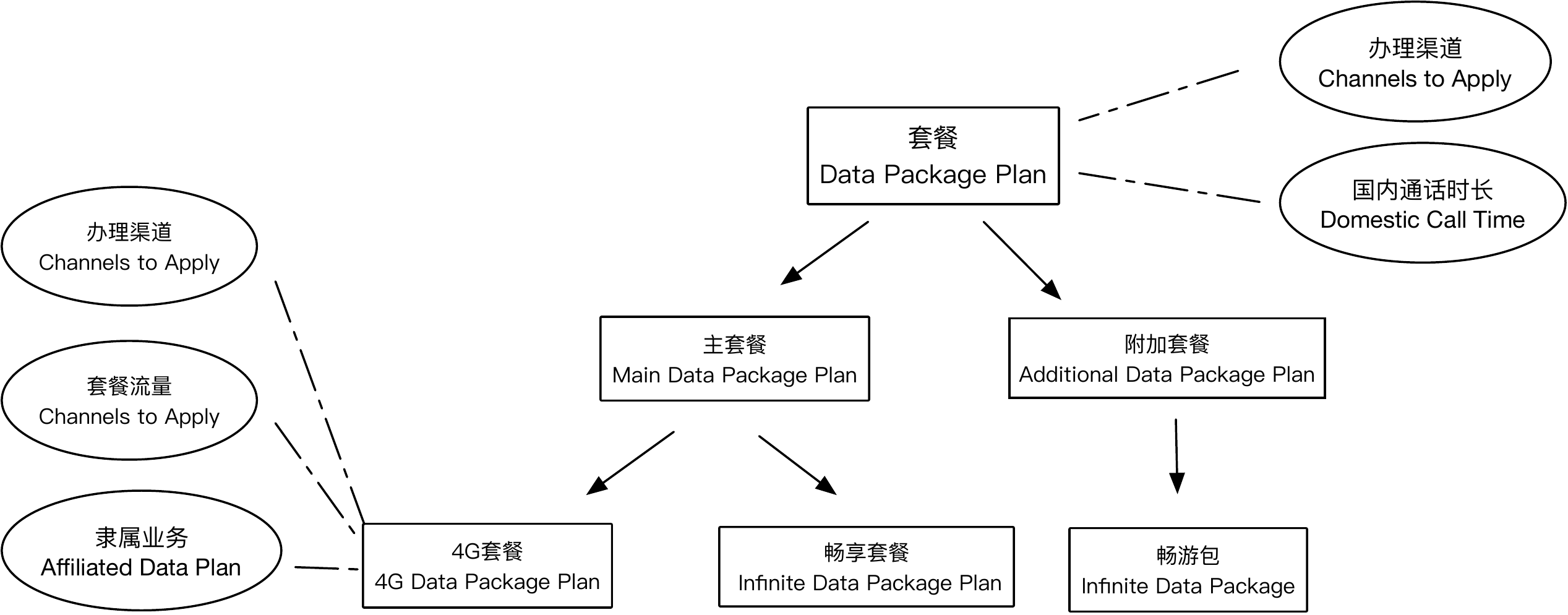}
    \caption{An illustrative example of a schema for the ``data package plan'' domain, with concepts (in rectangles) and attributes (in circles).}
    \label{fig:schema}
\end{figure}

\subsection{Track 1: Information Extraction from Dialog Transcripts}
\subsubsection{Schema}
The domain of a task-oriented dialogue system is
often characterized by an ontology, or say, a schema.
A \emph{schema} is a collection of hierarchical \emph{concepts} with \emph{attributes}, which is used to organize and interpret information in the domain. 
A illustrative schema including concepts and their attributes is shown in Figure~\ref{fig:schema}.
\emph{Entities} are instances of concepts.
Thus, all entities belonging to a concept have the attributes of the corresponding concept.
Attributes are also often called \emph{slots}.
The schema used in our annotation of the MobileCS dataset is shown in Figure~\ref{fig:schema-real}.

\subsubsection{Motivation}


In a task-oriented dialog system, after dialog state tracking, the system needs to query a task-related knowledge base (KB). The query result is important for the system to decide action and generate response. For system development from scratch in many real-world tasks, the knowledge base is often not readily available for training TOD systems.
Traditionally, expert labors are needed to construct the knowledge base.

\begin{figure}
    \centering
    \includegraphics[width=0.6\linewidth]{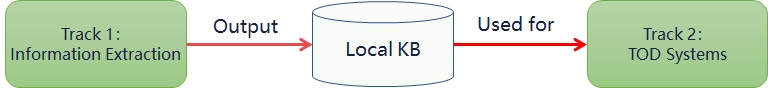}
    \caption{The connection between the two tracks in the sereTOD challenge.}
    \label{fig:connect}
\end{figure}

Given a mix of labeled and unlabeled dialog transcripts, Track 1 examines the task of training information extraction models to construct the local knowledge base for each dialog, which will be needed in training TOD systems in Track 2.
Ideally, we need a \emph{global KB}, which covers and fuses all public knowledge and all personal information in the domain.
But such a global KB is often difficult to obtain during the research phase.
Thus, in this challenge, we avoid this difficulty by considering a \emph{local KB} for each dialog.
The knowledge base is local in the sense that the mentioned entities with their mentioned attributes are extracted across all turns in a dialog, but there is no information fusion between dialogs\footnote{We leave information fusion across dialogs for future study.}.
A local KB for a dialog could be viewed as being composed of the relevant snapshots from the global KB.
With such local knowledge bases, we will still be able to drive the training of the TOD system.
Once the TOD system is trained in such a manner, the resulting TOD system potentially can work with a global KB.
In this challenge, for a first pilot study, the teams in Track 2 are only required to build TOD systems with local KBs.
The connection between Track 1 and Track 2 is illustrated in Figure \ref{fig:connect}.

\subsubsection{Task Definition}
Based on the schema, we define two sub-tasks for Track 1.

\begin{CJK*}{UTF8}{gbsn}
\paragraph{Entity extraction} This sub-task is to extract entity mentions with their corresponding \emph{entity-types (i.e., concepts)}, according to the set of entity-types defined in the schema.
In real-life dialogs, an entity may be mentioned in different surface forms. For example, ``50元流量包'' (50 Yuan data package plan) may have a number of different mentions in a multi-turn dialog: ``50元那个业务'' (50 Chinese Yuan plan), ``那个流量包'' (that package plan), ``刚才那个业务'' (that plan). 
Thus, entity extraction for the MobileCS dataset is more challenging than classic named entity recognition tasks (e.g., extracting person names), due to the informal, verbalized and loose form of the customer-service dialogs.

\paragraph{Slot filling} This sub-task is to extract \emph{slot-values} for entity \emph{slots (i.e., attributes)}. A set of slots is defined for each entity-type in the schema. For example, in utterance ``10GB套餐业务每月的费用是50块钱。'' (The price for 10GB data package plan is 50 Chinese Yuan per month), ``每月的费用是50块钱'' (50 Chinese Yuan per month) will be used to extract the value ``50块钱'' (50 Chinese Yuan) for the monthly price slot. An entity may have several mentions in a dialog, and the slots and values for an entity may scatter in multi-turn dialogs.
Thus, the task of slot filling requires \emph{entity resolution} and the assignment of the extracted slot-value pairs to the corresponding entity.
After entity extraction and slot filling, a local knowledge base (KB) will be constructed with all extracted entities with their attributes for each dialog. 
\end{CJK*}



\subsubsection{Evaluation}
Given a dialog in testing, the trained information extraction model is used to extract entities together with their slot-values.
We will evaluate and rank the submitted models by the extraction performance on test set. The evaluation metrics are based on {Precision}, {Recall} and {F1}. 
\begin{itemize}
    \item For entity extraction, the F1 is calculated at entity mention level: an entity mention is extracted correctly if and only if the mention span of the entity is labeled as the corresponding entity-type (i.e., concept). For entity extraction, the participants need to submit all the predicted mentions with their types.
    \item For slot filling, the F1 is calculated at triple level: an \emph{entity-slot-value triple} is extracted correctly if and only if 1) the mention span of the slot value is labeled as the corresponding slot type. 2) the slot-value pair is correctly assigned to the corresponding entity. For slot filling, the participants need to submit the extracted entities with entity resolution.
Each extracted entity may contain multiple mentions and is represented as a set of entity-slot-value triples. 
The performance of slot filling is measured by finding the best match between the extracted entities and the golden labeled entities using the Hungarian Algorithm\footnote{\url{https://en.wikipedia.org/wiki/Hungarian_algorithm}} and calculating the F1.
\end{itemize}

The average F1 scores of entity extraction and slot filling will be the ranking basis on leaderboard.
We will provide the following scripts and tools for the participants: 1) Baseline models for both sub-tasks; 2) Evaluation scripts to calculate the metrics.

\subsection{Track 2: Task-Oriented Dialog Systems}
\subsubsection{Motivation}
Most existing TOD systems require not only large amounts of annotations of dialog states and dialog acts (if used), but also a global knowledge base (KB) that covers all public knowledge and all personal information in the domain. Compared with previous work, the task in Track 2 has two main characteristics:
\begin{enumerate}
    \item There is no global KB but only a local KB (as shown in Listing~\ref{fig:goal-kb-example}) for each dialog, representing the unique information for each user, e.g., the user's package plan and remaining phone charges.
    \item Only a proportion of the dialogs is annotated with intents and local KBs. The teams are encouraged to utilize a mix of labeled and unlabeled dialogs to build a TOD system.
\end{enumerate}

\subsubsection{Task Definition}
The basic task for the TOD system is, for each dialog turn, given the dialog history, the user utterance and the local KB, to predict the user intent, query the local KB and generate appropriate system intent and response according to the queried information. 
For every labeled dialog, the annotations consist of user intents, system intents and a local KB. The local KB is obtained by collecting the entities and triples annotated for Track 1.
For unlabeled dialogs, there are no such annotations.


\subsubsection{Connection between Track 1 and Track 2}
As shown in Figure \ref{fig:connect},  the output from Track 1 is used as the local KB for Track 2.
Thus, the local KBs of unlabeled dialogs can be constructed by applying the information extraction model from Track 1 to extract entities and triples. 
For every unlabeled dialog in training, the organizers will provide extracted user information by running the baseline of Track 1, which the teams can use as the local KBs. 
The teams are allowed and encouraged to use their own information extraction models, built in Track 1, to construct the local KBs for training TOD systems in Track 2.





\subsubsection{Evaluation}
\paragraph{User Goal} \begin{CJK*}{UTF8}{gbsn}The main purpose of the TOD system is to fulfill the user's goal, such as querying data traffic and opening packages. The validation set and test set will contain annotations of user goals to facilitate the final evaluation. For each dialog, we accumulate the annotated user acts and triples mentioned by the user in  all the turns to obtain the user goal. User goals are in the form of lists. Each item in the list corresponds to an entity mentioned by the user, which contains both the information informed by the user and the attributes requested by the user. An example of user goal is shown in Listing~\ref{fig:goal-kb-example}. "?" denotes that the attribute "业务费用" is requested by the user, while other values denote that those attributes are informed by the user. The last key "意图" denotes the user intent for this entity. 
\begin{listing}
\begin{minipage}[b]{0.48\linewidth}
\begin{minted}[frame=lines,
               framesep=2mm,
               linenos=false,
               xleftmargin=5pt,
               baselinestretch=0.8,
               %fontsize=\footnotesize,
               tabsize=2]{python}
[
  {
    "name":"套餐",
    "type":"主套餐",
    "通话时长":"二百七十分钟",
    "业务费用":"六十八块钱"
  }
  
]

\end{minted}
\end{minipage}
\hfill
\begin{minipage}[b]{0.48\linewidth}
\begin{minted}[frame=lines,
               framesep=2mm,
               linenos=false,
               xleftmargin=5pt,
               baselinestretch=0.8,
               %fontsize=\footnotesize,
               tabsize=2]{python}
[
  {
    "name":"套餐",
    "type":"主套餐",
    "通话时长":"二百七十分钟",
    "业务费用":"?",
    "意图":["求助-查询"]
  }
]
\end{minted}
\end{minipage}
\caption{Examples of local KB (left) and user goal (right). The English version can be seen in Listing~\ref{fig:goal-kb-example-eng} in Appendix~\ref{sec:eng_example}.} 
\label{fig:goal-kb-example}
\end{listing}
\end{CJK*}

\paragraph{Automatic Evaluation}
In order to measure the performance of TOD systems, both automatic evaluation and human evaluation will be conducted. 
For automatic evaluation, metrics include Precision/Recall/F1 score, Success rate and BLEU score.  P/R/F1 are calculated for both predicted user intents and system intents.
Success rate is the percentage of generated dialogs that achieve user goals. BLEU score evaluates the fluency of generated responses. 
The combined score in Track2 are calculated as follows: Combined score = User intent F1 + System intent F1 + Success + BLEU/50.

\paragraph{Human Evaluation}
We will perform human evaluation for different TOD systems, where real users interact with those systems according to randomly given goals. 
For each dialog, the user will score the system on a 5-point scale (1-5) by the following three metrics. The higher, the better.
\begin{itemize}
    \item Success. This metric measures if the system successfully completes the user goal by interacting with the user;
    \item Coherency. This metric measures whether the system’s response is logically coherent with the dialogue context;
    \item Fluency. The metric measures the fluency of the system’s response.
\end{itemize}


The average scores from automatic evaluation and human evaluation will be the main ranking basis on leaderboard.
We will provide the following scripts and tools for the participants: 1) A baseline system; 2) Evaluation scripts to calculate the corpus-based metrics.

\section{The MobileCS Dataset}
\subsection{Structure of Annotations}
\begin{figure*}
    \centering
    \includegraphics[width=1.0\linewidth]{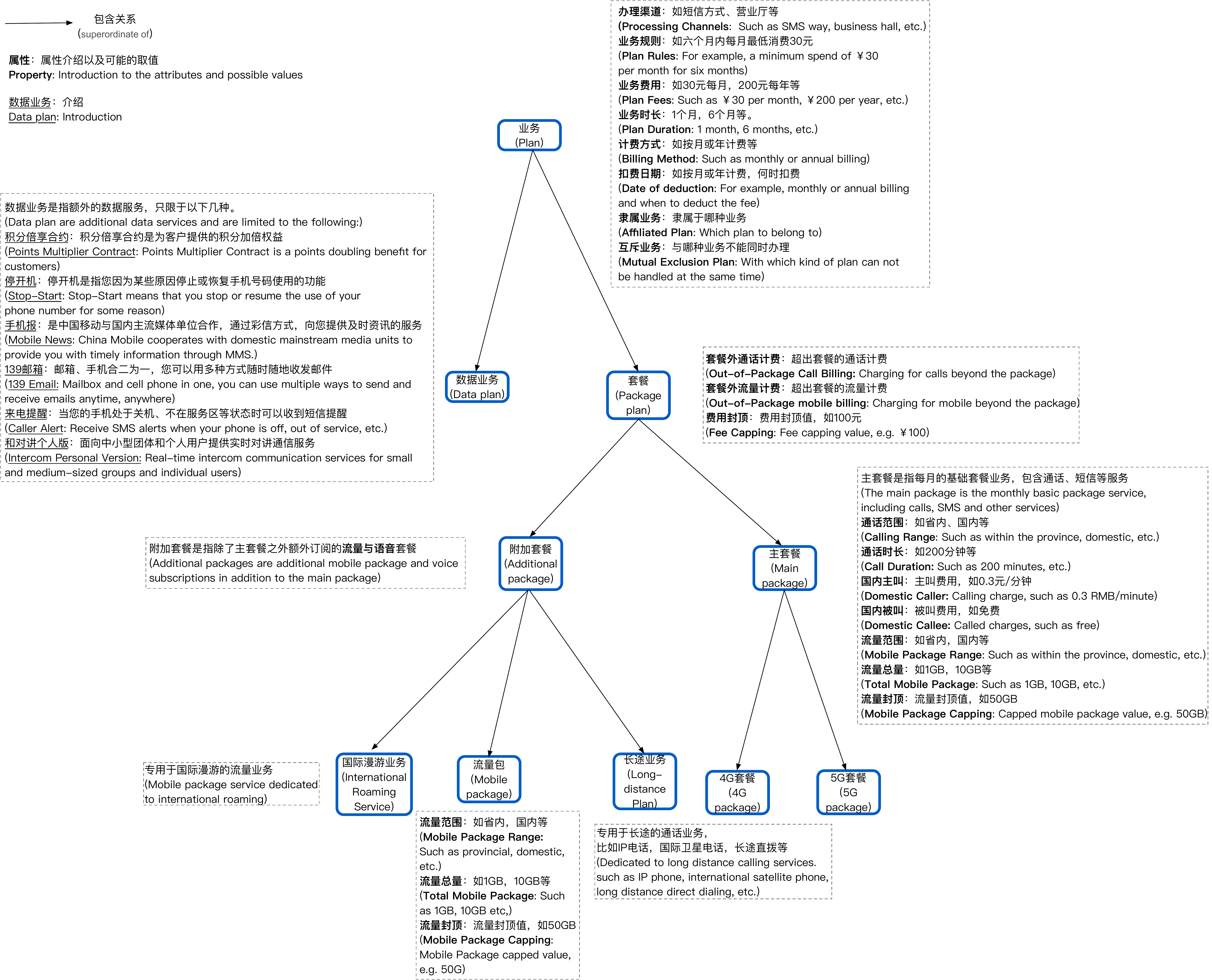}
    \caption{The schema for annotating the MobileCS dataset, including the concepts (in blue boxes) and attributes (in dotted boxes).
The attribute "业务规则" (service rule) for the concept "业务" (service) may cover a wide range of information. 
    For the concept "数据业务" (data service), only the six entities underlined are considered.
    The concept "用户" (user) with its attributes is shown in Figure \ref{fig:user}.}
    \label{fig:schema-real}
\end{figure*}

The MobileCS dataset is annotated according to its schema and intent-sets.
The schema used in our annotation of the MobileCS dataset is shown in Figure~\ref{fig:schema-real}, which is needed for both information extraction and TOD systems.
The intent-sets for users and custom-service staffs are shown in Listing \ref{fig:intent-sets}, which are mainly required for building TOD systems.
The annotations for a dialog consist of entities, entity attributes, attribute values (i.e., slot-values), user intents, and customer-service intents that appear in each turn. The meanings of these terms are introduced as follows:
\begin{CJK*}{UTF8}{gbsn}
\begin{itemize}
    \item Entities: refer to instances of concepts in the schema. Entities in the MobileCS dataset are mostly related to telecommunication services, such as “King Package”.
    \item Entity attributes: refer to the attributes of entities, such as “业务费用” (service cost). 
    Besides, user’s personal information is also annotated, such as the attribute “用户状态” (user status) for the concept “用户” (user), as shown in Figure \ref{fig:user}.
    \item Attribute values: refer to the values of attributes. For example, the attribute value of “业务费用” (service cost) is “30元” (30 yuan).
    \item Intents: at each turn, for the user and the customer-service separately, one or more intents from the corresponding intent-set are labeled.
\end{itemize}
\end{CJK*}

\begin{CJK*}{UTF8}{gbsn}
\end{CJK*}




\begin{CJK*}{UTF8}{gbsn}
\begin{listing}
\begin{minted}[frame=lines,
               framesep=2mm,
               linenos=false,
               xleftmargin=5pt,
               xrightmargin=5pt,
               baselinestretch=0.8,
               %fontsize=\footnotesize,
               tabsize=2]{python}

  "用户意图":{
    "求助-查询":"让客服帮忙查询流量套餐等相关信息",
    "求助-故障":"让客服帮忙解决各种故障", 
    "提供信息":"向客服陈述相关信息", 
    "投诉反馈":"向客服投诉，表达不满", 
    "取消":"取消某个套餐或者活动", 
    "询问":"求助-查询以外的普通询问", 
    "请求重复":"请求对方重复所说的话",  
    "主动确认":"向对方主动确认信息，一般会给出陈述", 
    "被动确认":"被动确认信息，一般只给出几个表示肯定的字", 
    "否认":"否认对方提出确认问题", 
    "问候":"打招呼，一般出现在第一轮对话",
    "再见":"表示想结束对话的意图都算作再见", 
    "客套":"客套话，包括谢谢，不客气，没关系等等", 
    "其他":"不属于上面任何一种意图"
},
"客服意图":{
    "通知":"通知用户相关信息，包括查询到的用户流量套餐信息以及客服所采取的操作等",
    "建议":"建议用户采取某项措施，比如重启手机", 
    "询问":"询问用户相关信息", 
    "引导":"引导用户继续发言", 
    "请求重复":"请求对方重复所说的话", 
    "主动确认":"向对方主动确认信息，一般会给出陈述", 
    "被动确认":"被动确认信息，一般只给出几个表示肯定的字", 
    "否认":"否认对方提出确认问题", 
    "问候":"打招呼，一般出现在第一轮对话", 
    "抱歉":"向用户道歉",
    "再见":"表示想结束对话的意图都算作再见", 
    "客套":"客套话，包括谢谢，不客气，没关系等等", 
    "其他":"不属于上面任何一种意图"
}

\end{minted}
\caption{The intent-sets for users and custom-service staffs, respectively. The English version can be seen in Listing~\ref{fig:intent-sets-eng} in Appendix~\ref{sec:eng_example}.} 
\label{fig:intent-sets}
\end{listing}

\begin{listing}
\begin{minted}[frame=lines,
               framesep=2mm,
               linenos=false,
               xleftmargin=5pt,
               baselinestretch=0.8,
               %fontsize=\footnotesize,
               tabsize=2]{python}
{
    "[SPEAKER 1]": "哦，请讲",
    "[SPEAKER 2]": "问一下，我这个，我这个卡开通的什么业务啊"
},
{
    "[SPEAKER 1]": "哦，您目前用的是，十八元的这个基础套餐",
    "[SPEAKER 2]": "哦，我想问一下，就是我，我当时开那个，呃开通了一个呃"
},
{
    "[SPEAKER 1]": "那个活动，没有",
    "[SPEAKER 2]": "他给我说了半年，呃，半年现在一直扣着我的钱呀"
},
{
    "[SPEAKER 1]": "它是三十元上网免费用三个月的，到四月十六号这个活动就结束\
                    了然后加一块钱这个活动，要求三十元上网，半年不能关",
    "[SPEAKER 2]": "当时他跟我说的是，半年的时候是免费使用，由他开通的"
}
\end{minted}
\caption{A dialog example before annotation. The English version can be seen in Listing~\ref{fig:before-eng} in Appendix~\ref{sec:eng_example}.} 
\label{fig:before}
\end{listing}
\end{CJK*}

\subsubsection{Annotation File Format} 
The annotations are saved in json file format.
A dialog example before annotation is shown in Listing \ref{fig:before}. Each turn in a dialog is annotated with three types of labels:
\begin{itemize}
    \item entities (“ents”);
    \item (entity, attribute, attribute value)-triples (“triples”);
    \item user and customer-service intents.
\end{itemize}
Next, we introduce the specific guidelines for annotating these three types of labels, which should be helpful for understanding and using these labels.

\subsection{Guideline for Annotations}

\subsubsection{Entity Annotation} 

\begin{CJK*}{UTF8}{gbsn}
At each turn, the entities and entity-types (along with their positions) are annotated, whether they appeared in user utterances or custom-service utterances.
For example, for the utterance: “哦，您目前用的是，十八元的这个基础套餐”, the labels are: entity-name “十八元的这个基础套餐”, entity-type “套餐”, and the position information of mentioned entity.

\begin{itemize}
    \item Entity-name. Entity-name is the literal name of the entity mentioned in the utterance. In our annotation, the description information is also contained in the entity-name. For example, for “十八元的基本套餐” appeared in a certain utterance, the entity-name is “十八元的基本套餐”. 
    A more example is: for “十八元的这个基础套餐”, the entity-name is “十八元的这个基础套餐”.
    \item Entity-ID. The same entity may be mentioned by different expressions in a dialogue. For example, “和风套餐” may be mentioned by a number of expressions such as “这个套餐”, “和风”, “那个套餐”, and these different expressions actually refer to the same entity. 
    Thus, entity-ID in the form of “ent-xx” are used in annotating entities, where xx represents the specific ID. Different expressions of the same entity are labeled as the same ID, and different entities are labeled with different IDs. For a multi-turn dialog, the ID starts from 1, and +1 when encountering a new entity. See the example in Listing \ref{fig:entity-ID}.
    
   \begin{CJK*}{UTF8}{gbsn}
\begin{listing}
\begin{minted}[frame=lines,
               framesep=2mm,
               linenos=false,
               xleftmargin=5pt,
               baselinestretch=0.8,
               %fontsize=\footnotesize,
               tabsize=2]{python}

{
    "[SPEAKER 1]": "你详细说一下那个和风套餐。",
    "[SPEAKER 2]": "和风套餐提供了很多服务，每个月20GB流量。",
    "info": {
        "ents": [
            {
                "name": "和风套餐", 
                "id": "ent-1", 
                "type": "套餐", 
                "pos": [[1, 8, 12],[2, 0, 4]]
            }
        ],
        "triples": [
            {
                "ent-id": "ent-1",
                "ent-name": "和风套餐",
                "prop": "流量总量",
                "value": "每个月20GB"
            }
        ]
    }
},
{
    "[SPEAKER 1]": "这个套餐多少钱？",
    "[SPEAKER 2]": "38元一个月。",
    "info": {
        "ents": [
            {
                "name": "套餐", 
                "id": "ent-1",
                "type": "套餐", 
                "pos": [1, 2, 4]
            }
        ],
        "triples": [
            {
                "ent-id": "ent-1"
                "ent-name": "套餐",
                "prop": "业务费用",
                "value": "38元一个月"
            }
        ]
    }
},
{
    "[SPEAKER 1]": "还有其他套餐吗？",
    "[SPEAKER 2]": "有的，比如大王套餐等。",
    "info": {
        "ents": [
            {
                "name": "大王套餐",
                "id": "ent-2",
                "type": "套餐",
                "pos": [2, 5, 9]
            }
        ],
        "triples": []
    }
}

\end{minted}
\caption{An example of annotating entity-IDs. The English version can be seen in Listing~\ref{fig:entity-ID-eng} in Appendix~\ref{sec:eng_example}.} 
\label{fig:entity-ID}
\end{listing}

\end{CJK*}
    
    \item Entity-type. 
    When an entity is mentioned in an utterance, its entity-type is annotated, according to the schema in Figure \ref{fig:schema-real}.
    Note that the schema contains a hierarchical collection of entity-types, which could be viewed as a rooted tree. The further from the root, the finer-grained the entity-type.
    When choosing entity-types from the schema, the most fine-grained entity-type is used for labeling.
    For example, consider that the “套餐” type contains “主套餐”. So when the utterance clearly states that the entity is “主套餐”, it is labeled as the type of “主套餐”, rather than the type of “套餐”. 
    \item Entity-position. Entity-position is labeled as a triple. The first field denotes whether the entity mention appears in the first or the second utterance in the turn.
    The second and third fields denote the beginning and ending positions of the mention in the utterance, which is called a \emph{span}. 
    The positions are numbered from 0.
    The ending position is set to the position of the last symbol in the span plus 1.
    Note that punctuation symbols and spaces are also counted.
    See the example in Listing \ref{fig:entity-ID}.
    
\end{itemize}

\subsubsection{Attribute Annotation} 
\begin{figure}
    \centering
    \includegraphics[width=0.5\linewidth]{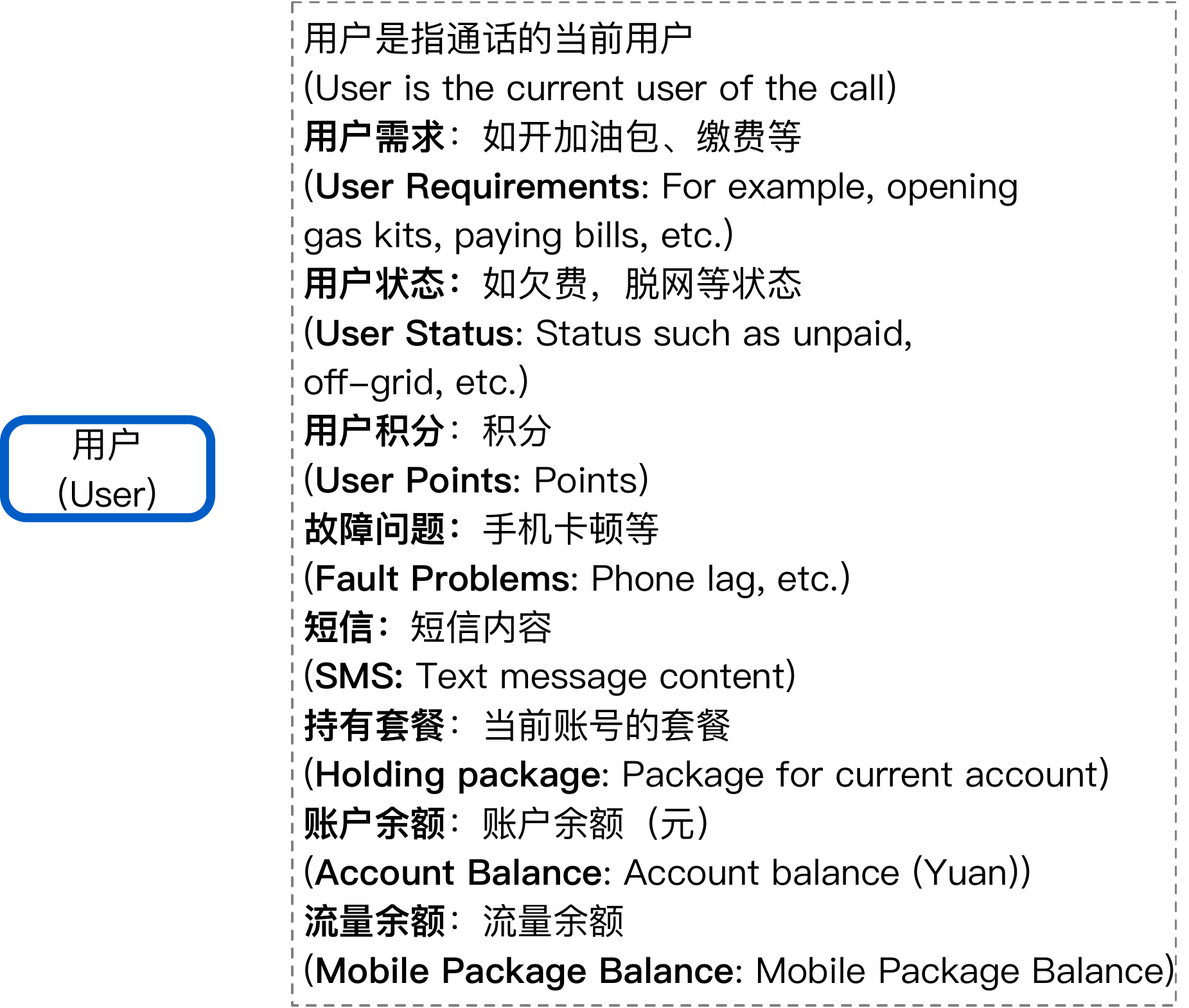}
    \caption{User personal information, i.e., the concept "用户" (user) with its attributes.}
    \label{fig:user}
\end{figure}

At each turn, entity attributes and their corresponding attribute values are annotated in the form of "triples".
See the format of annotating attributes in Listing \ref{fig:attribute-format}. 
There may have multiple attribute values in a turn, so “triples” is a list.
Remarkably, there are two classes of concepts in the schema for the MobileCS dataset, i.e., services and users, which are shown in Figure \ref{fig:user} and Figure \ref{fig:schema-real} respectively.
That means that the user in a dialog is a special entity, and so its attribute annotation is different from annotating attributes for services.

\paragraph{Annotating attributes for services} 
In annotation, the attribute value is firstly located in a certain turn, e.g., “38元一个月”. Then, the attribute corresponding to the attribute value needs to be identified from the schema, which is “业务费用” in this example. 
Next, the entity, which the attribute value belongs to, is identified, which may appear in the current turn or not. There are two cases.
\begin{itemize}
    \item The entity appears in the current turn. Then, the ID of the corresponding entity and the “name” of the entity can be labeled in the current turn. 
    \item The entity does not appear in current turn, but is mentioned in some previous turn. Then, the ID of the corresponding entity is labeled, and the entity-name is annotated as “NA”. See an example in Listing \ref{fig:attribute_case2}.
\end{itemize}


\begin{CJK*}{UTF8}{gbsn}
\begin{listing}
\begin{minted}[frame=lines,
               framesep=2mm,
               linenos=false,
               xleftmargin=5pt,
               baselinestretch=0.8,
               %fontsize=\footnotesize,
               tabsize=2]{python}

"triples": [
    {
        "ent-id": <ent-id>, // 实体的id
        "ent-name": <ent-name>, //实体在本轮对话中的表达
        "prop": <属性>,  // 属性(property)
        "value": <属性值>, // 对话中出现的属性值(property value)
    }
]

\end{minted}
\caption{The format of annotating attributes} 
\label{fig:attribute-format}
\end{listing}

\end{CJK*}
   
\begin{CJK*}{UTF8}{gbsn}
\begin{listing}
\begin{minted}[frame=lines,
               framesep=2mm,
               linenos=false,
               xleftmargin=5pt,
               baselinestretch=0.8,
               %fontsize=\footnotesize,
               tabsize=2]{python}

{
    "[SPEAKER 1]": "你详细说一下那个和风套餐。",
    "[SPEAKER 2]": "和风套餐提供了很多服务，每个月20GB流量。",
    "info": {
        "ents": [
            {
                "name": "和风套餐", 
                "id": "ent-1", 
                "type": "套餐", 
                "pos": [[1, 8, 12], [2, 0, 4]]
            }
        ],
        "triples": [
            {
                "ent-id": "ent-1",
                "ent-name": "和风套餐",
                "prop": "流量总量",
                "value": "每个月20GB"
            }
        ]
    }
},
{
    "[SPEAKER 1]": "多少钱？",
    "[SPEAKER 2]": "38元一个月。",
    "info": {
        "ents": [],
        "triples": [
            {
                "ent-id": "ent-1",
                "ent-name": "NA",
                "prop": "业务费用",
                "value": "38元一个月"
            }
        ]
    }
}

\end{minted}
\caption{A special case in annotating attributes: the entity does not appear in the current turn, where the attribute value appears.
The entity-name is labeled as “NA”.} 
\label{fig:attribute_case2}
\end{listing}

\end{CJK*} 

\paragraph{Annotating attributes for users}     
The user in a dialog is an entity from the concept "用户" (user), which has it own attributes (i.e., the user's personal information), as shown in Figure \ref{fig:user}.
Since we use local KBs, the current user is the only entity from the concept "用户" (user) in each dialog.
Therefore, there is no need to annotate this entity.
In annotating attributes for the user, "ent-id" and "ent-name" are annotated as "NA".
See the example in Listing \ref{fig:personal_info}.


  
\begin{CJK*}{UTF8}{gbsn}
\begin{listing}
\begin{minted}[frame=lines,
               framesep=2mm,
               linenos=false,
               xleftmargin=5pt,
               baselinestretch=0.8,
               %fontsize=\footnotesize,
               tabsize=2]{python}

{
    "[SPEAKER 1]": "这边显示您的状态属于脱网状态。",
    "[SPEAKER 2]": "啊是吗？",
    "info": {
        "ents": [],
        "triples": [
            {
                "ent-id": "NA",
                "ent-name": "NA",
                "prop": "用户状态",
                "value": "脱网"
            }
        ]
    }
}

\end{minted}
\caption{An example of annotating attributes for user personal information. The English version can be seen in Listing~\ref{fig:personal_info-eng} in Appendix~\ref{sec:eng_example}.} 
\label{fig:personal_info}
\end{listing}

\end{CJK*}

\subsubsection{Intent Annotation} 
At each turn, intent annotation is conducted for the user and the customer-service separately.
The intent-sets for users and custom-service staffs are shown in Listing \ref{fig:intent-sets}。
One or more intents from the user intent-set are labeled for the current user utterance. Intent labeling is similar for the current custom-service utterance, but is based on the custom-service intent-set.
Multiple intents are separated by commas. The intent annotation cannot be empty. Intents that are difficult to determine are annotated as “其他”.

When the user or the customer-service staff expresses the intent to obtain some entity information (such as “求助-查询”, “询问”, “主动确认”, etc.), more annotations are appended to the annotated intents. There are three cases, which are exemplified respectively in Listing \ref{fig:intent-example}.
\begin{itemize}
    \item When a user wants to obtain information about an entity, the corresponding \emph{entity-ID and attribute} are appended to the annotated intent, and are connected with "-". 
    In the first example in Listing \ref{fig:intent-example}, “求助-查询（ent-1-业务规则）” indicates that the user wants to query the business rule of the entity ent-1 (三十八元套餐).
    \item When a user wants to obtain the user information as shown in Figure \ref{fig:user}, the corresponding \emph{attribute} is appended to the the annotated intent.
    In the second example in Listing \ref{fig:intent-example}, “求助-查询（账户余额）” means that the user wants to know the balance of his/her account.
    \item When some queries do not involve any particular entities, for example, the user wants wants the customer-service to recommend some services to him/her, the corresponding \emph{entity-type} is appended to the the annotated intent.
       In the third example in Listing \ref{fig:intent-example}, “求助-查询（主套餐）” means that the user wants to query which main packages are available.
\end{itemize}


\begin{CJK*}{UTF8}{gbsn}
\begin{listing}
\begin{minted}[frame=lines,
               framesep=2mm,
               linenos=false,
               xleftmargin=5pt,
               baselinestretch=0.8,
               %fontsize=\footnotesize,
               tabsize=2]{python}

"样例1":{
    "[SPEAKER 1]": "诶",
    "[SPEAKER 2]": "就是我这个三十八的这个套餐，这到因为我当时记得签约了，\
                    是到什么时候，你能看到吗",
    "客服意图":"引导",
    "用户意图":"求助-查询（ent-1-业务规则）"
    },
"样例2":{
    "[SPEAKER 1]": "但它这个话费还没有把这个月租冲掉的哈",
    "[SPEAKER 2]": "我知道，我现在我的话费还有多少",
    "客服意图":"通知",
    "用户意图":"求助-查询（账户余额）"
    },
"样例3":{
    "[SPEAKER 1]": "哦，请讲",
    "[SPEAKER 2]": "问一下，我这个，我这个卡开通的什么业务啊",
    "客服意图":"引导",
    "用户意图":"求助-查询（业务）"
    }

\end{minted}
\caption{An example of intent annotation. Annotations for entities and triples are omitted for simplicity. The English version can be seen in Listing~\ref{fig:intent-example-eng} in Appendix~\ref{sec:eng_example}.} 
\label{fig:intent-example}
\end{listing}

\end{CJK*} 
 
Note that in a multi-turn dialog, the customer-service staff may speak first or the user may speak first. In annotation, the labeling order of the user intent and the customer-service intent is determined according to the speaker's order.

\section{Challenge Rules}
\begin{itemize}
\item The challenge website is \url{http://seretod.org/Challenge.html}. Teams should submit the registration form to \url{seretod2022@gmail.com}, which will be reviewed by the organizers. 
\item Teams are required to sign an Agreement for Challenge Participation and Data Usage. Data will be provided to approved teams.
\item For teams that participate in Track 1, the scores will be ranked according to the performance for Track 1. The teams can choose to participate only in Track 1.
\item For teams that participate in Track 2, they can use the baseline system provided by the organizers or use the system developed by themselves for Track 1. The ranking is based on the performance for Track 2.
\item Participants are allowed to use any external datasets, resources or pre-trained models which are publicly available.
\item Participants are NOT allowed to do any manual examination or modification of the test data.
\end{itemize}

\end{CJK*}

\bibliography{ref}
\bibliographystyle{iclr2022_conference}
\newpage
\appendix
\section{Appendix}
\subsection{Examples in English}
\label{sec:eng_example}
\begin{listing}
\begin{minipage}[b]{0.48\linewidth}
\begin{minted}[frame=lines,
               framesep=2mm,
               linenos=false,
               xleftmargin=5pt,
               baselinestretch=0.8,
               %fontsize=\footnotesize,
               tabsize=2]{python}
[
  {
    "name":"Package",
    "type":"Main package",
    "call minutes":"270 minutes",
    "fee":"Sixty eight yuan"
  }
  
]

\end{minted}
\end{minipage}
\hfill
\begin{minipage}[b]{0.48\linewidth}
\begin{minted}[frame=lines,
               framesep=2mm,
               linenos=false,
               xleftmargin=5pt,
               baselinestretch=0.8,
               %fontsize=\footnotesize,
               tabsize=2]{python}
[
  {
    "name":"Package",
    "type":"Main package",
    "call minutes":"270 minutes",
    "fee":"?",
    "intent":["query"]
  }
]
\end{minted}
\end{minipage}
\caption{Examples of local KB (left) and user goal (right) in English.} 
\label{fig:goal-kb-example-eng}
\end{listing}

\begin{listing}
\begin{minted}[frame=lines,
               framesep=2mm,
               linenos=false,
               xleftmargin=5pt,
               xrightmargin=5pt,
               baselinestretch=1.0,
               %fontsize=\footnotesize,
               tabsize=2]{python}

  "User intent":{
    "query":"Ask the system to help query package related information",
    "failure":"Ask the system to help solve various failure", 
    "inform":"State relevant information to the system", 
    "feedback":"Complain to the system", 
    "cancel":"Cancel a package or activity", 
    "ask":"Ordinary inquiry other than query", 
    "pardon":"Ask the other party to repeat",  
    "active confirm":"Actively confirm the information to the other party,\ 
                      and generally give a statement", 
    "passive confirm":"Passively confirm information and generally only\
                       give a few words indicating yes", 
    "deny":"Deny the confirmation question", 
    "greet":"Greetings usually appear in the first turn of dialogue",
    "bye":"Intention to end the conversation", 
    "politeness":"Polite words, including thank you, you're welcome, etc", 
    "others":"It does not belong to any of the above intentions"
},
"System intent":{
    "inform":"Inform users of relevant information, such as the queried \
              user package information",
    "suggest":"Suggest the user to take a certain measure", 
    "ask":"Ask users for relevant information", 
    "guide":"Guide the user to continue speaking", 
    "pardon":"Ask the other party to repeat",  
    "active confirm":"Actively confirm the information to the other party, \
                      and generally give a statement", 
    "passive confirm":"Passively confirm information and generally only\
                       give a few words indicating yes", 
    "deny":"Deny the confirmation question", 
    "greet":"Greetings usually appear in the first turn of dialogue",
    "sorry":"sorry",
    "greet":"Greetings usually appear in the first turn of dialogue",
    "bye":"Intention to end the conversation", 
    "politeness":"Polite words, including thank you, you're welcome, etc", 
    "others":"It does not belong to any of the above intentions"
}

\end{minted}
\caption{The intent-sets for users and custom-service staffs in English, respectively.} 
\label{fig:intent-sets-eng}
\end{listing}

\begin{listing}
\begin{minted}[frame=lines,
               framesep=2mm,
               linenos=false,
               xleftmargin=5pt,
               baselinestretch=1.0,
               %fontsize=\footnotesize,
               tabsize=2]{python}
{
    "[SPEAKER 1]": "Oh, please",
    "[SPEAKER 2]": "I want to ask, what service does my, my card open"
},
{
    "[SPEAKER 1]": "Oh, you are currently using this basic package of 18 yuan",
    "[SPEAKER 2]": "Oh, I want to ask that I opened that, er, er"
},
{
    "[SPEAKER 1]": "That activity? no",
    "[SPEAKER 2]": "He told me it's for half a year, er, it has been \
                    withholding my money for half a year now"
},
{
    "[SPEAKER 1]": "It is free to use the Internet for three months at 30 yuan.\
                    The activity will be over by April 16, and then you need to\
                    add one yuan for the activity. It requires that the Internet\
                    be used for 30 yuan and cannot be closed for half a year",
    "[SPEAKER 2]": "At that time, he told me that it was free to use for half \
                    a year, and he opened it"
}
\end{minted}
\caption{A dialog example before annotation in English.} 
\label{fig:before-eng}
\end{listing}

\begin{listing}
\begin{minted}[frame=lines,
               framesep=2mm,
               linenos=false,
               xleftmargin=5pt,
               baselinestretch=0.8,
               %fontsize=\footnotesize,
               tabsize=2]{python}

{
    "[SPEAKER 1]": "Please elaborate on the HeFeng package",
    "[SPEAKER 2]": "HeFeng package provides many services, with 20GB\
                    traffic per month.",
    "info": {
        "ents": [
            {
                "name": "HeFeng package", 
                "id": "ent-1", 
                "type": "package", 
                "pos": [[1, 8, 12],[2, 0, 4]]
            }
        ],
        "triples": [
            {
                "ent-id": "ent-1",
                "ent-name": "HeFeng package",
                "prop": "Total data traffic",
                "value": "20GB traffic per month"
            }
        ]
    }
},
{
    "[SPEAKER 1]": "How much is this package?",
    "[SPEAKER 2]": "38 yuan a month.",
    "info": {
        "ents": [
            {
                "name": "Pacakge", 
                "id": "ent-1",
                "type": "package", 
                "pos": [1, 2, 4]
            }
        ],
        "triples": [
            {
                "ent-id": "ent-1"
                "ent-name": "Package",
                "prop": "fee",
                "value": "38 yuan a month"
            }
        ]
    }
},
{
    "[SPEAKER 1]": "Are there any other packages?",
    "[SPEAKER 2]": "Yes, such as Dawang package",
    "info": {
        "ents": [
            {
                "name": "Dawang package",
                "id": "ent-2",
                "type": "package",
                "pos": [2, 5, 9]
            }
        ],
        "triples": []
    }
}

\end{minted}
\caption{An example of annotating entity-IDs in English.} 
\label{fig:entity-ID-eng}
\end{listing}

\begin{listing}
\begin{minted}[frame=lines,
               framesep=2mm,
               linenos=false,
               xleftmargin=5pt,
               baselinestretch=0.8,
               %fontsize=\footnotesize,
               tabsize=2]{python}

{
    "[SPEAKER 1]": "Your status here is offline",
    "[SPEAKER 2]": "Ah, really?",
    "info": {
        "ents": [],
        "triples": [
            {
                "ent-id": "NA",
                "ent-name": "NA",
                "prop": "User status",
                "value": "offline"
            }
        ]
    }
}

\end{minted}
\caption{An example of annotating attributes for user personal information in English.} 
\label{fig:personal_info-eng}
\end{listing}

\begin{listing}
\begin{minted}[frame=lines,
               framesep=2mm,
               linenos=false,
               xleftmargin=5pt,
               baselinestretch=1.0,
               %fontsize=\footnotesize,
               tabsize=2]{python}

"Example1":{
    "[SPEAKER 1]": "Hey",
    "[SPEAKER 2]": "It's my thirty-eight package, I remember signing it\ 
                    at that time can you see when it expires?",
    "System intent":"guide",
    "User intent":"query(ent-1-rules)"
    },
"Example2":{
    "[SPEAKER 1]": "But the monthly rent hasn't been written off yet",
    "[SPEAKER 2]": "I know. How much do I have left in my phone now",
    "System intent":"inform",
    "User intent":"query(Account balance)"
    },
"Example3":{
    "[SPEAKER 1]": "Oh, please",
    "[SPEAKER 2]": "I want to ask, what service does my, my card open",
    "System intent":"guide",
    "User intent":"query(service)"
    }

\end{minted}
\caption{An example of intent annotation in English. Annotations for entities and triples are omitted for simplicity.} 
\label{fig:intent-example-eng}
\end{listing}
\end{document}